\documentclass[]{TEAI}

\usepackage{helvet}
\usepackage{bbding}
\usepackage{siunitx}

\usepackage{amsmath,amsfonts,bm}

\def\eqref#1{equation~\ref{#1}}

\def\1{\bm{1}}

\DeclareMathAlphabet{\mathsfit}{\encodingdefault}{\sfdefault}{m}{sl}
\SetMathAlphabet{\mathsfit}{bold}{\encodingdefault}{\sfdefault}{bx}{n}

\title{Daily-Omni: Towards Audio-Visual Reasoning with Temporal Alignment across Modalities}

\author[1]{Ziwei Zhou}
\author[1]{Rui Wang}
\author[1]{Zuxuan Wu}
\author[1]{Yu-Gang Jiang}

\affiliation[1]{Fudan University}

\checkdata[Website]{\url{https://github.com/Lliar-liar/Daily-Omni}}
\abstract{Recent Multimodal Large Language Models (MLLMs) achieve promising performance on visual and audio benchmarks independently. However, the ability of these models to process cross-modal information synchronously remains largely unexplored. We introduce \textbf{Daily-Omni}, a multiple-choice Audio-Visual QA benchmark featuring \textbf{684} real-world videos and \textbf{1,197} questions spanning 6 task families that explicitly require cross-modal temporal reasoning. To support scalable benchmark construction, we develop a semi-automatic pipeline for annotation, cross-modal consistency refinement, temporal alignment elicitation, and text-only leakage filtering, followed by human verification. We further provide a diagnostic evaluation suite and extensively evaluate \textbf{24} foundation models under \textbf{37} model--modality settings (Audio+Video / Audio-only / Video-only / Text-only). Finally, we include a \textbf{training-free modular diagnostic baseline} that composes off-the-shelf unimodal models to serve as a diagnostic baseline and to illustrate how explicit temporal alignment signals affect performance. Results indicate that many end-to-end MLLMs still struggle on alignment-critical questions, suggesting that robust cross-modal temporal alignment remains an important open challenge.}

\begin{document}
\maketitle

\section{Introduction}

In our daily lives, non-textual modalities like vision and audio convey much richer information than text alone. They are essential for us to understand and interact with the physical world. Therefore, advancements in Multimodal Large Language Models (MLLMs) capable of comprehensively understanding the multi-modal information are a crucial foundation for achieving artificial intelligence capable of interacting with the physical world and observing the outcome of its operations. 

Recent MLLMs \citep{Qwen3-VL,Qwen3-Omni,xu_qwen25-omni_2025,tang_enhancing_2024,noauthor_introducing_2024,zhang_internlm-xcomposer25-omnilive_2024,fu_vita-15_2025,liu_ola_2025,cheng_videollama_2024,li_baichuan-omni-15_2025,xie_mini-omni2_2024,guo_m2-omni_2025,team_reka_2024,microsoft2025phi4minitechnicalreportcompact,lu2023unifiedio2scalingautoregressive,liu_nexus-o_2025,gemmateam2025gemma3technicalreport,gemini25} have demonstrated groundbreaking capabilities spanning audio (ASR, sound classification, captioning) and visual domains (OCR, VQA, video grounding), with significant accuracy improvements over previous benchmarks. However, existing methods still face several limitations. Firstly, many MLLMs predominantly focus on visual abilities, often neglecting the importance of other modalities like audio. This oversight may stem from the fact that current visual datasets are more abundant, of higher quality, and cover a broader range of tasks compared to audio datasets. Existing audio datasets \citep{Librispeech,wang2020covost2massivelymultilingual,poria2019meldmultimodalmultipartydataset,chen_vggsound_2020, Audioset,Vocalsound} tend to prioritize speech-related or music-related tasks and basic sound classification, often overlooking more complex yet crucial tasks such as reasoning over generic sounds. Consequently, many MLLMs incorporate only speech encoders as their primary auditory component or rely heavily on speech-related datasets for audio pre-training. This architectural limitation fundamentally restricts their ability to comprehend rich acoustic environments where non-speech sounds (e.g., environmental noises, mechanical failures, or emotional cues in non-verbal vocalizations) carry critical semantic information. Secondly, the current landscape lacks high-quality multimodal datasets that integrate temporally aligned auditory and visual information. Existing audio-visual datasets and benchmarks \citep{yun_pano-avqa_2021,li2022learninganswerquestionsdynamic,avqa,li_omnibench_2024,hong_worldsense_2025,gong_av-odyssey_2024,sungbin2025avhbenchcrossmodalhallucinationbenchmark,geng2025longvalevisionaudiolanguageeventbenchmarktimeaware,omnivideobench} suffer from three persistent limitations. First, several focus on specialized scenarios \citep{yun_pano-avqa_2021,li2022learninganswerquestionsdynamic} such as musical performances or panoramic environments, thereby introducing domain-specific biases. 
\begin{figure}[h!]
  \centering
  \includegraphics[width=1\textwidth]{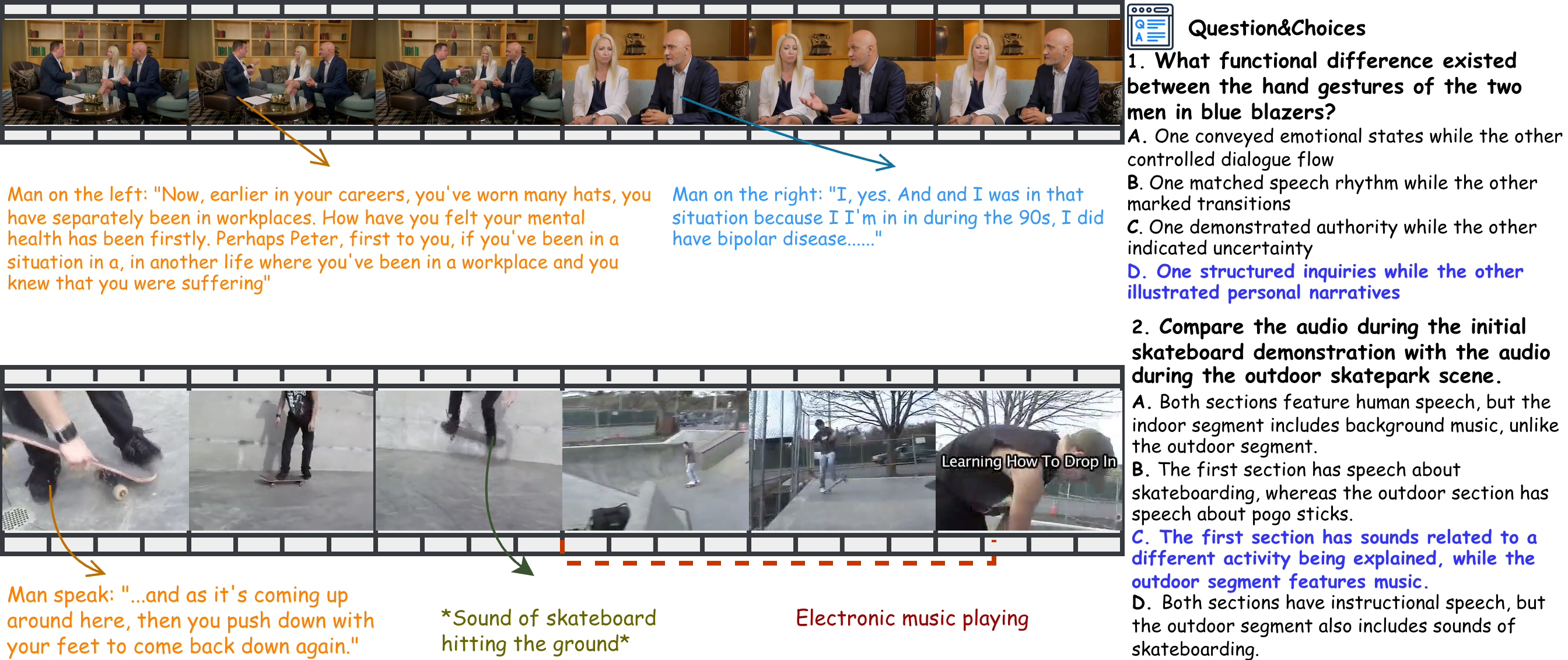}
  \caption{\textbf{Illustrative examples from the Daily-Omni benchmark}. These cases highlight the dataset's focus on complex cross-modal reasoning. The top example requires temporally aligning specific speakers' voices with their gestures, while the bottom example requires comparing acoustic environments across different video segments.}
  \label{Fig:qa_example}

\end{figure}
Second, many employ static image-audio pairs \citep{li_omnibench_2024,gong_av-odyssey_2024} that disregard crucial temporal dynamics inherent in real-world video contexts. Third, current tasks are often too narrow, with many benchmarks focusing on specific applications such as captioning or open-ended responses \citep{geng2025longvalevisionaudiolanguageeventbenchmarktimeaware}. The common absence of rigorous evaluation frameworks and standardized metrics makes it hard to reliably compare results. While WorldSense \citep{hong_worldsense_2025} established a valuable audio-visual multi-choice QA benchmark for daily scenarios through its meticulous dataset curation, two critical limitations persist: (1) it lacks a systematic framework for scalable QA generation, relying instead on manual annotation processes that hinder dataset expansion; and (2) the benchmark serves primarily as an evaluation tool, offering limited methodological guidance on enhancing model capabilities via explicit training protocols or architectural modifications. This dual limitation constrains both the benchmark's adaptability to emerging domains and its practical utility in driving model improvements.

In this paper, we introduce \textbf{Daily-Omni}, an Audio-Visual Questioning and Answering benchmark designed to evaluate temporally-aligned multimodal reasoning. The videos are sampled from diverse datasets \citep{Audioset,fu2024videommefirstevercomprehensiveevaluation,Farre2024FineVideo} and segmented into 30-second or 60-second intervals to systematically assess model performance across different temporal contexts. Our core contributions are three-fold:
\begin{itemize}
    \item \textbf{The Daily-Omni Benchmark}: We construct a high-quality dataset comprising 684 real-life videos rich in both audio and visual dynamics, featuring 1197 multiple-choice QA pairs across 6 major tasks ranging from fundamental audio-visual event alignment to complicated cross-modal reasoning.
    \item \textbf{Scalable QA Generation Pipeline}: We propose an automated framework encompassing video annotation, revision, audio-visual temporal alignment, QA generation, and optimization. This highly efficient pipeline significantly reduces human labor, requiring only 30 hours for a single annotator to complete the quality filtering process (achieving an approximate 30\% acceptance rate), thereby ensuring the benchmark's scalability.
    \item \textbf{Extensive Evaluation \& Diagnostic Protocol}. We benchmark 24 representative foundation models under 37 model--modality settings (AV / V-only / A-only / text-only) and provide diagnostic analyses including modality ablations and alignment-sensitivity. We additionally include a training-free modular reference baseline (Daily-Omni Agent) composed of off-the-shelf unimodal models, intended as a sanity-check reference.
\end{itemize}

Our comprehensive results reveal an important phenomenon: many current end-to-end MLLMs struggle with tasks requiring deep audio-visual temporal integration, to the extent that our simple, explicit-alignment baseline outperforms several recent open-source omni-modal models. This extensive evaluation demonstrates that robust temporal alignment mechanisms are still insufficient or missing in many unified architectures, underscoring both the necessity of our benchmark and a clear direction for future multimodal reasoning research.

\section{Related Works}
\subsection{Multimodal Large Language Models}
Recent advancements in Multimodal Large Language Models (MLLMs) primarily fall into three categories: Audio Language Models (ALMs) incorporating audio capabilities \citep{tang_salmonn_2024,gong2024listenthinkunderstand,chu_qwen-audio_2023,chu2024qwen2audiotechnicalreport,ghosh2025audioflamingo2audiolanguage,ghosh2024gamalargeaudiolanguagemodel,goel2025audioflamingo3advancing}, Visual Language Models (VLMs) adding visual understanding \citep{liu2025nvilaefficientfrontiervisual,bai_qwen25-vl_2025,lu2024deepseekvlrealworldvisionlanguageunderstanding,wang2024emu3nexttokenpredictionneed}, and Omni-modal Language Models (OLMs) combining both audio and visual modalities \citep{noauthor_introducing_2024,li_baichuan-omni-15_2025,zhang_internlm-xcomposer25-omnilive_2024,team_reka_2024,cheng_videollama_2024,liu_ola_2025,microsoft2025phi4minitechnicalreportcompact,fu_vita-15_2025,guo_m2-omni_2025,Qwen3-Omni,xu_qwen25-omni_2025,liu_nexus-o_2025,gemmateam2025gemma3technicalreport}. These MLLMs often employ a modular architecture, utilizing separate encoders for audio and visual inputs. In the audio domain, several works \citep{chen2022beatsaudiopretrainingacoustic,li2024mertacousticmusicunderstanding,radford2022robustspeechrecognitionlargescale} proposed audio encoders to extract sound representations. Some models \citep{tang_salmonn_2024,liu_ola_2025,zhang_internlm-xcomposer25-omnilive_2024} even integrate multiple audio encoders specialized for different sound types (e.g., speech, music). Similarly, visual encoders \citep{li2023blip2bootstrappinglanguageimagepretraining,dai2023instructblipgeneralpurposevisionlanguagemodels,liu2023visualinstructiontuning,dehghani2023patchnpacknavit} are used to process images and videos. However, this modular approach struggles to capture the crucial temporal correlations inherent in synchronized audio-visual streams. This limitation arises because audio and visual inputs are encoded independently. Although multi-modal positional embedding techniques like TMRoPE \citep{xu_qwen25-omni_2025} enhance cross-modal temporal understanding to some extent, effective methods for temporally aligning multimodal data remain relatively scarce. Additionally, in several MLLMs \citep{microsoft2025phi4minitechnicalreportcompact,fu_vita-15_2025,zhang_internlm-xcomposer25-omnilive_2024} , the audio encoder primarily serves to process user instructions---akin to how the text modality is used---rather than perceiving the environment.

\begin{table}[htbp]
\centering
\caption{\textbf{Comparison of audio-visual benchmarks.} We detail publication, modality (A: audio, V: video, I: image), size, and question type (MC: multiple choice, DF: defined word, BB: bounding boxes). 'Automatic Scalability' refers to automated expansion methods. 'Open-Domain' and 'General Sound' denote genre and sound diversity.}
\scriptsize
\setlength{\tabcolsep}{2.5pt}

\resizebox{\linewidth}{!}{
\begin{tabular}{l | c c c c c c c}
\toprule
\textbf{Benchmarks} & \textbf{Pub} & \textbf{Modality} & \textbf{\#QA} & \textbf{Question Type} & \textbf{Automatic Scalability} & \textbf{Open-Domain} & \textbf{General Sound} \\
\midrule
AVQA  &ACM MM'22 & V+A   &57,335 & MC &\XSolidBrush & \Checkmark & \XSolidBrush  \\
Music-AVQA &CVPR'22 & V+A  & 45,867& DF & \XSolidBrush &\XSolidBrush &\Checkmark\\
Pano-AVQA  &ICCV'21 & V+A & 51,700& DF \& BB & \XSolidBrush &\XSolidBrush &\XSolidBrush\\
OmniBench &ARXIV'24 & I+A  & 1,142 & MC & \XSolidBrush & \Checkmark & \Checkmark \\
AV-Odyssey &ARXIV'24 & I+A   &4,555& MC & \XSolidBrush & \Checkmark & \Checkmark\\
WorldSense &ICLR'26  & V+A  & 3,172& MC & \XSolidBrush & \Checkmark & \Checkmark\\
\midrule
\textbf{Daily-Omni}& --& V+A& 1197& MC & \Checkmark& \Checkmark& \Checkmark \\
\bottomrule
\end{tabular}
}
\label{tab:dataset_comparison} 

\end{table}

\subsection{Audio-Visual Understanding Datasets and Benchmarks}
The development of uni-modal datasets and associated tasks for audio and vision has driven advancements in Audio Language Models (ALMs) and Visual Language Models (VLMs). Visual datasets and benchmarks \citep{li2024llavanextinterleavetacklingmultiimagevideo,fu2024videommefirstevercomprehensiveevaluation,wu_star_2024,mangalam_egoschema_2023,liu2023visualinstructiontuning,wu2024longvideobench,zhang2024videoinstructiontuningsynthetic,yue2024mmmumassivemultidisciplinemultimodal,yue2024mmmuprorobustmultidisciplinemultimodal,li2023seedbenchbenchmarkingmultimodalllms,fu2024ocrbenchv2improvedbenchmark,gao2017talltemporalactivitylocalization,hu2025videommmuevaluatingknowledgeacquisition,omnivideobench} primarily focus on tasks for static images (OCR, grounding, segmentation, classification, and question-answering) and dynamic videos (captioning, temporal grounding, and understanding), while audio datasets and benchmarks \citep{ghosh2025audioflamingo2audiolanguage,Audioset,chen_vggsound_2020,Librispeech,yang2024airbenchbenchmarkinglargeaudiolanguage} address speech-related tasks (ASR, emotion recognition, and entity recognition) and non-speech tasks (such as sound classification and audio grounding). However, while attempts at creating audio-visual datasets date back to at least 2021 \citep{yun_pano-avqa_2021}, these early efforts often had significant limitations. For example, Music-AVQA \citep{li2022learninganswerquestionsdynamic} focused specifically on music performance videos, while Pano-AVQA \citep{yun_pano-avqa_2021} centered on panoramic videos. Others, such as AVQA \citep{avqa} and OmniBench \citep{li_omnibench_2024}, were restricted to short, simple videos or static images. Furthermore, AV-Odyssey \citep{gong_av-odyssey_2024} heavily emphasized specific audio tasks, such as recognizing timbre and loudness, rather than broader audio-visual understanding. While WorldSense \citep{hong_worldsense_2025}, a concurrent work, also provides a valuable benchmark for real-world audio-visual question-answering, the efficient scalability of AVQA datasets and the enhancement of OLM abilities still require further exploration.

\section{Daily-Omni}

This section describes the construction of Daily-Omni, including data curation, video annotation, and QA synthesis and verification. We also describe an evaluation protocol and include a training-free modular reference baseline to help interpret model behaviors. Daily-Omni contains \textbf{684} real-world videos from all \textbf{11} YouTube categories and \textbf{1,197} multiple-choice questions spanning \textbf{6} task families. The dataset includes \textbf{550} questions on 60-second clips and \textbf{647} questions on 30-second clips. Table~\ref{tab:dataset_comparison} compares Daily-Omni with prior audio-visual benchmarks.

\begin{figure}[htbp]
    \centering
    \begin{subfigure}[b]{0.43\linewidth}
        \centering
        \includegraphics[width=\linewidth]{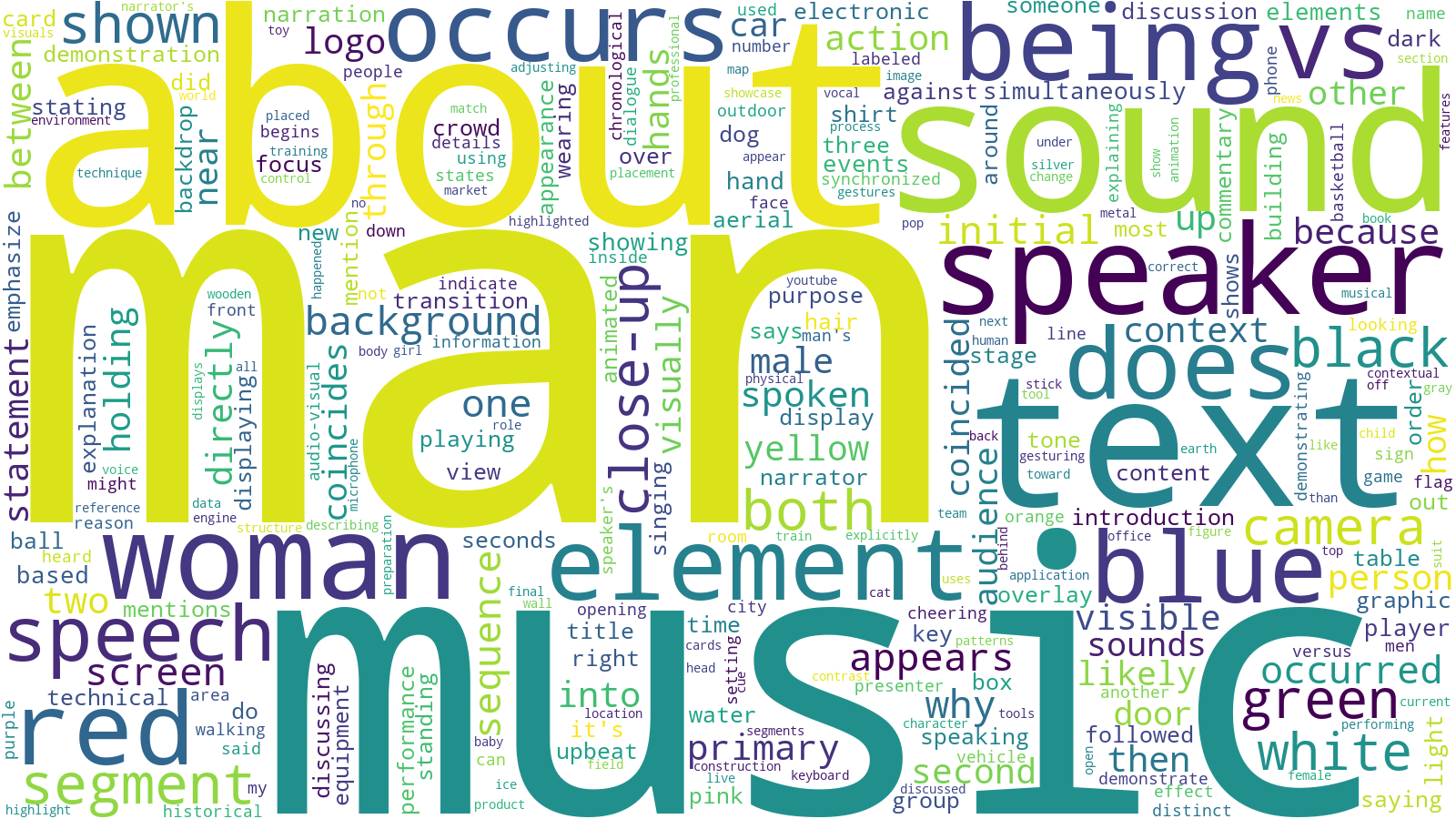}
        \caption{Word cloud of Daily-Omni QA pairs}
        \label{Fig:sub_wordcloud}
    \end{subfigure}
    \hfill 
    \begin{subfigure}[b]{0.55\linewidth}
        \centering
        \includegraphics[width=\linewidth]{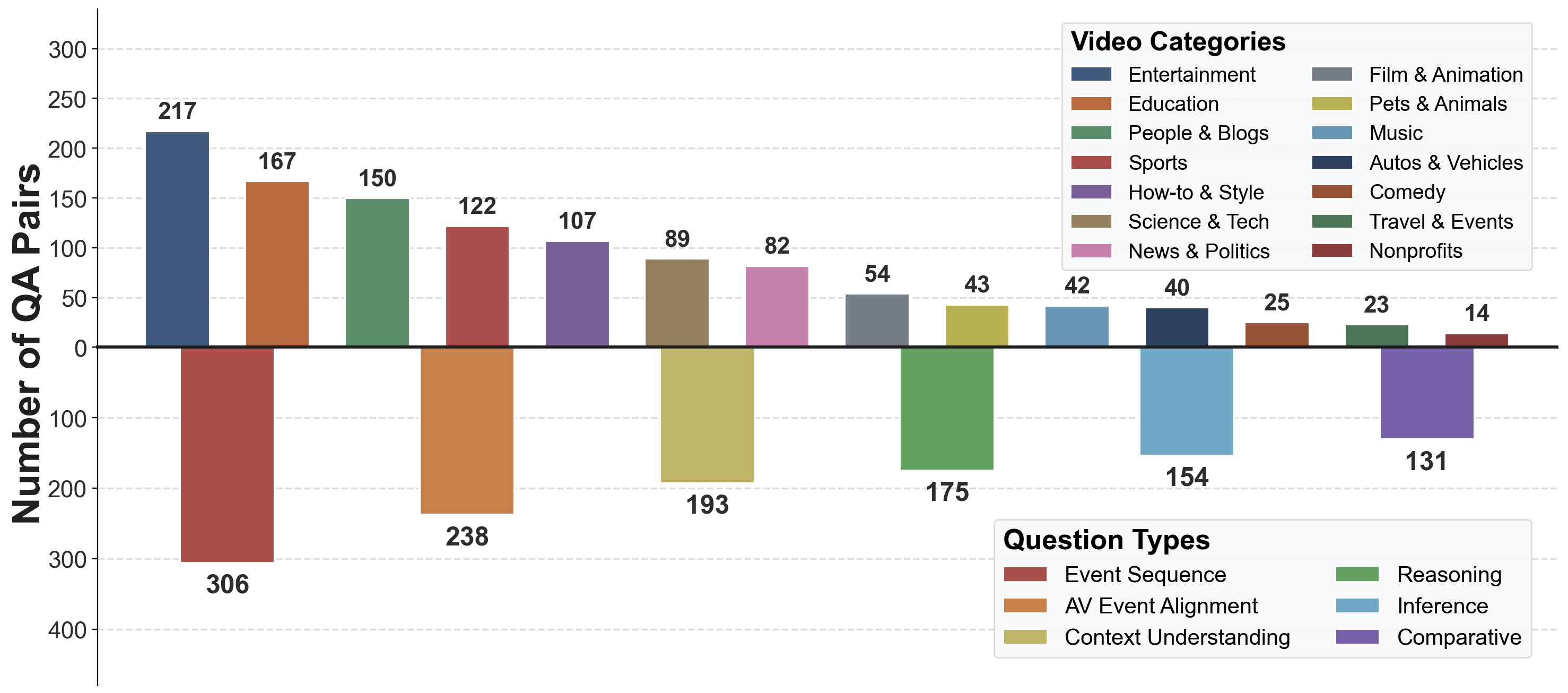}
        \caption{Statistics and distributions of the benchmark}
        \label{Fig:sub_qa_count}
    \end{subfigure}
    
    \caption{\textbf{Overview of the Daily-Omni benchmark.} (a) The prominent vocabulary in the word cloud highlights our focus on rich human activities, diverse acoustic events, and temporal reasoning. (b) The bar chart details the long-tail distribution of the 14 source video categories (top) and the composition of the 6 core question types (bottom).}
    \label{Fig:dataset_overview}
\end{figure}

\subsection{Data Curation}
We curate videos with rich temporal dynamics in vision and diverse everyday sounds beyond speech (e.g., music and environmental events). We exclude largely static talking-head videos and non-English videos to reduce confounds from multilingual content. Figure \ref{Fig:dataset_overview}
shows the word cloud of questions and choices in Daily-Omni and the distribution of video categories and QA types.
Our videos are sourced from AudioSet~\citep{Audioset}, Video-MME~\citep{fu2024videommefirstevercomprehensiveevaluation}, and FineVideo~\citep{Farre2024FineVideo}. For AudioSet, we retrieve the original YouTube videos and extract 30s/60s segments that contain the target sound events. We use Whisper-V3-Large~\citep{radford2022robustspeechrecognitionlargescale} to ensure spoken content is present. For Video-MME and FineVideo, we prioritize clips with substantial audio information and salient temporal visual changes to increase scenario diversity.

\subsection{Data Annotation \& QA Construction}
\begin{figure}[htbp]
  \centering
  \includegraphics[width=0.9\textwidth]{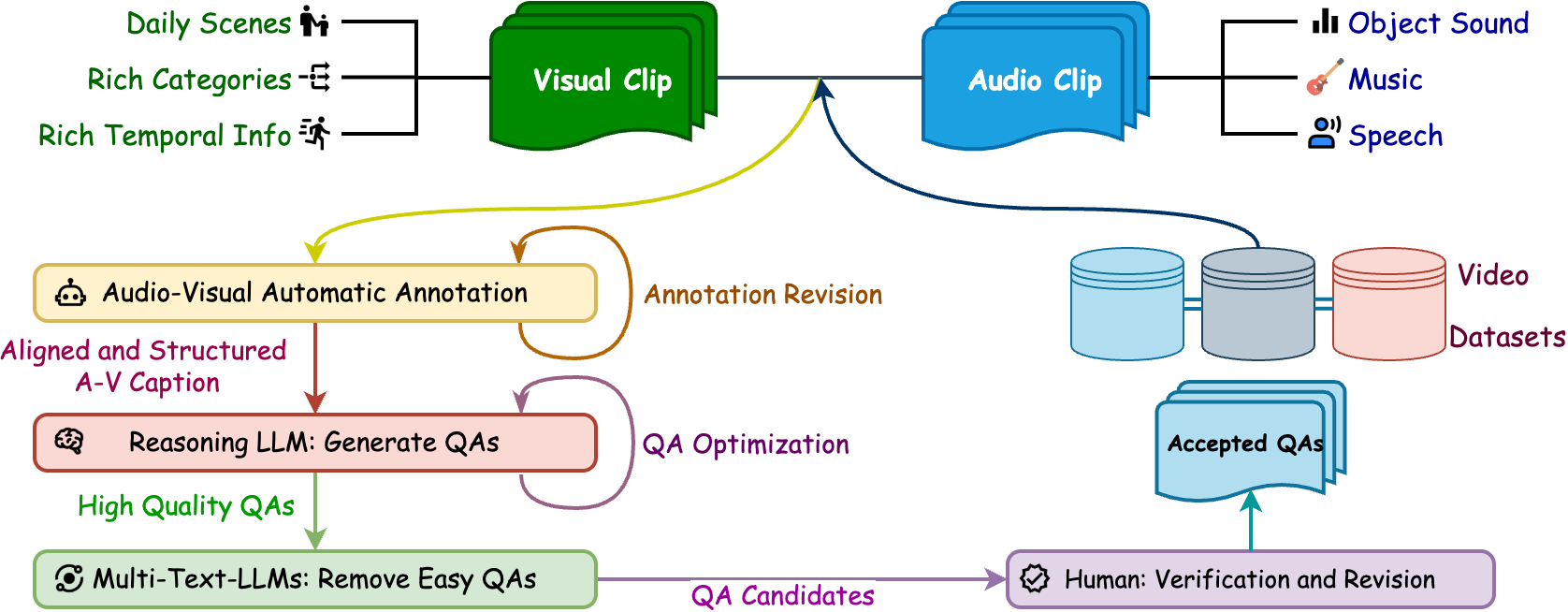}
  \caption{\textbf{The outline of Daily-Omni QA construction pipeline.} The arrows indicates the sequence of the processes.}
  \label{Fig:qa_pipeline}

\end{figure}

We developed a pipeline that employs MLLMs to generate and revise visual and audio annotations. Concurrently, Reasoning Large Language Models (LLMs) are utilized to construct and optimize the associated questions, choices, and answers. To obtain detailed annotations while ensuring cost-effectiveness, we specifically used Gemini 2.0 Flash \citep{noauthor_introducing_2024} for the annotation task and Deepseek-R1 \citep{deepseekai2025deepseekr1incentivizingreasoningcapability} for QA construction and optimization. \textbf{Figure \ref{Fig:qa_pipeline}} provides an outline of this process.

\textbf{Segment Annotation}  Recognizing that even state-of-the-art Multimodal Large Language Models (MLLMs) can exhibit cross-modal hallucinations \citep{sungbin2025avhbenchcrossmodalhallucinationbenchmark}, we employed Gemini 2.0 Flash to annotate the audio and visual modalities independently. Additionally, since MLLM performance is known to degrade when processing long audio clips, we segmented the videos prior to annotation. Specifically, each clip was divided into three equal, shorter segments (e.g., 10s segments for 30s clips, 20s segments for 60s clips) to improve the MLLM's subsequent audio annotation quality.

\textbf{Visual \& Audio Revision} After obtaining visual annotations from the video segments (processed without audio) and audio annotations from the corresponding audio segments, we use Gemini 2.0 Flash to perform a consistency check on the visual annotations. For this step, we prompt the model with the complete video clip, allowing it to review the segment annotations and ensure overall coherence. For example, referencing the full-length video, the model verifies whether a person described in an early segment is the same individual appearing in subsequent segments and generates a consistent revised annotation. Subsequently, audio annotations undergo refinement using a Reasoning LLM (Gemini 2.0 was employed in the initial phase of the project, with Deepseek-R1 used in later stages). This model leverages the consistent visual annotations to perform cross-modal correction, rectifying sound misidentifications and identifying sound sources. For example, if the audio is annotated as a 'generic impact sound' but the visual annotation for the same segment shows a 'door slamming shut', the Reasoning LLM uses this visual context to correct the audio description to 'door slamming sound' and attributes the sound's origin to the observed door.

\begin{figure}[htbp]
  \centering
  \includegraphics[width=1\textwidth]{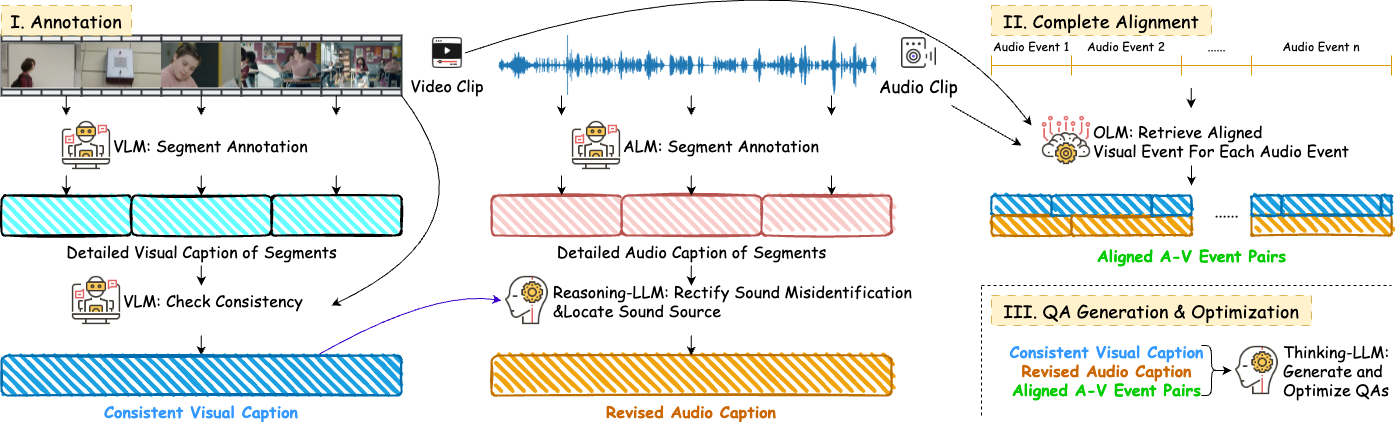}
  \caption{\textbf{Details of Daily-Omni annotation generation, revision and event alignment.} For cost-efficiency, we align all events with one query.}
  \label{Fig:annotation_detail}

\end{figure}

\textbf{Visual \& Audio Event Alignment} At this stage, we have sequences of visual and audio events annotated within consecutive 10s or 20s segments. However, these segment-level annotations do not explicitly specify the temporal alignment between individual visual and audio events, i.e., which specific events occurred simultaneously. To establish this precise cross-modal event concurrency, we proposed a \textbf{event aligning technique}. By prompting Gemini 2.0 Flash with the complete audio-visual clip, we instruct it to identify the visual event(s) occurring concurrently with each identified audio event. These aligned audio-visual event pairs provide sufficient information to infer the temporal relationships between any audio and visual event within the sequence. The details of annotation generation, revision and event alignment is shown in \textbf{Figure \ref{Fig:annotation_detail}}. We further assess the reliability of this alignment step via manual verification on a randomly sampled subset of 100 videos: over \textbf{90\%} of aligned audio--visual event pairs are judged correct.

\textbf{QA Construction} Using the consistent visual annotations, revised audio annotations, and aligned event pairs derived from each video, we prompted Deepseek-R1 to generate multi-choice questions covering the following types: (1) \textbf{AV Event Alignment}: To determine which audio and visual events occurred simultaneously with each other; (2) \textbf{Event Sequence}: To determine the temporal sequence of visual and audio events in the video; (3) \textbf{Reasoning}: To explain the cause or reason behind the occurrence of a visual or audio event in the video; (4) \textbf{Inference}: To speculate on information not explicitly presented in the video; (5) \textbf{Comparative}: To compare the similarity or difference between the audio and visual information of two or more events in the video; (6) \textbf{Context Understanding}: To determine the contextual information surrounding a specific event in the video.
Since Daily-Omni aims to evaluate MLLM perception and reasoning within real-life audio-visual scenarios, we deliberately exclude certain question types. Specifically, counting and measuring questions are omitted, as they often rely on a single modality rather than integrated multi-modal understanding. Similarly, purely knowledge-based queries such as celebrity identification are not included.

\textbf{QA Optimization and Quality Control} By avoiding strict question templates, the generated QAs exhibit greater creativity, incorporate complicated logic, and sometimes feature obscure descriptions, making them more closely resemble questions asked in real-life scenarios. However, the generated questions and choices can sometimes contain excessive textual information, potentially allowing powerful models to infer the correct answer using text alone, without engaging with the audio-visual content. Therefore, Deepseek-R1 was employed again to remove superfluous textual information from the questions and choices. Additionally, it replaced obviously incorrect options with more challenging distractors, thereby increasing the difficulty and reducing the potential for correct answers based on guessing alone. Subsequently, we evaluated the optimized questions using two powerful LLMs, GPT-4o \citep{openai_gpt-4o_2024} and Deepseek-V3 \citep{deepseekai2025deepseekv3technicalreport}, providing them with only the textual questions and choices (no audio-visual context). Questions that could be answered correctly by both LLMs under this text-only condition were discarded, as they did not necessitate multimodal reasoning. This automated filtering step resulted in approximately 47\% of the candidate QAs being discarded. Finally, the remaining QAs underwent manual evaluation for quality control. Human evaluators examined each QA, verifying: (1) that there was exactly one unambiguously correct answer among the choices, (2) that the proposed answer was indeed the correct one, and (3) that answering the question genuinely required comprehensive audio-visual capabilities. Based on this assessment, evaluators either accepted the QA for inclusion in the final benchmark or rejected it. Facilitated by the automated pipeline, the final human evaluation process was highly efficient. A \textbf{single annotator} used less than \textbf{30 hours} to review the candidates and establish the final set of 1197 QAs, corresponding to an acceptance rate of approximately \textbf{30\%} during this manual review stage.
\begin{figure}[htbp]

  \centering
  \includegraphics[width=0.95\textwidth]{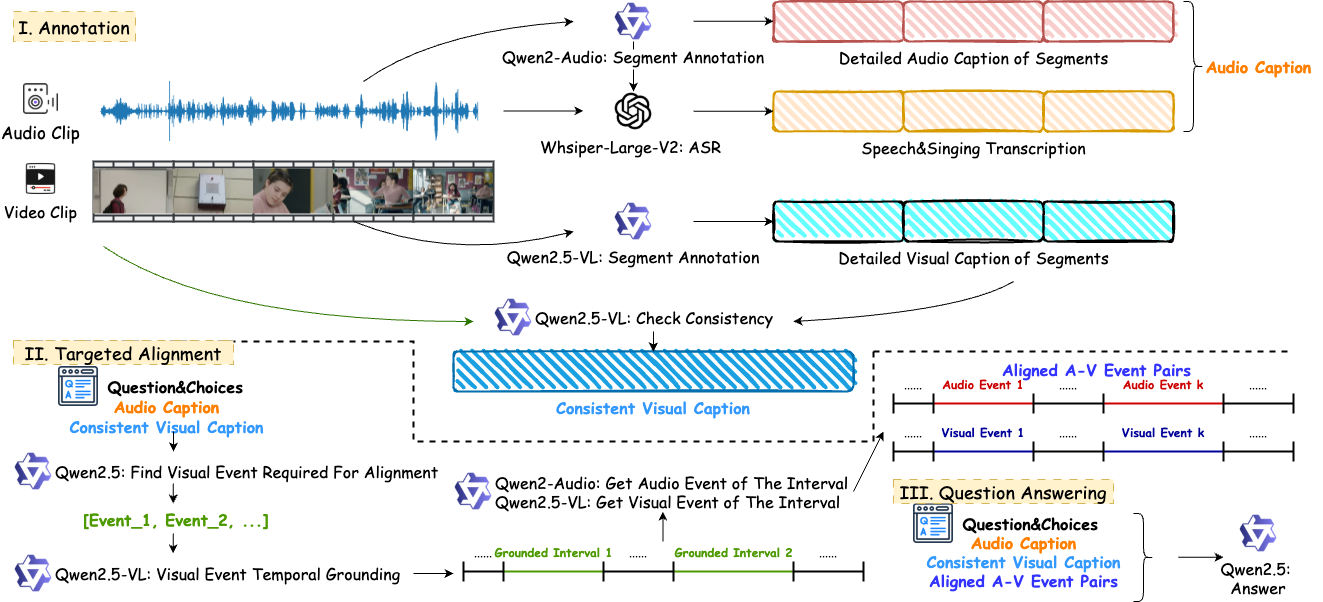}
  \caption{\textbf{The outline of Daily-Omni Agent workflow.}}
  \label{Fig:baseline_agent}

\end{figure}
\subsection{Diagnostic Baseline: Daily-Omni Agent}
To establish a heuristic reference for MLLMs and diagnose the importance of explicit temporal awareness in audio-visual question answering, we constructed a training-free diagnostic pipeline, termed the \textbf{"Daily-Omni Agent"}. This baseline, designed to understand audio-visual context and answer related questions, integrates several models: Qwen2-Audio \citep{chu2024qwen2audiotechnicalreport}, Qwen2.5-VL-7B \citep{bai_qwen25-vl_2025}, Whisper-Large-V2 \citep{radford2022robustspeechrecognitionlargescale}, and the text-based Qwen2.5-14B-Instruct \citep{qwen2025qwen25technicalreport}.
As shown in \textbf{Figure \ref{Fig:baseline_agent}}, when presented with a question, its choices, and the associated video context, the agent first divides both the video and audio streams into three segments of equal duration. Subsequently, it independently generates annotations for these segments: visual annotations using Qwen2.5-VL and audio annotations using Qwen2-Audio. Additionally, we use Whisper-Large-V2 to provide a transciption of speech and singing of each segment, as Qwen2-Audio tends to omit this in its annotations. Following the segment annotation, the agent utilizes Qwen2.5-VL to perform a \textbf{consistency check} on the visual annotations using the complete video, mirroring the revision process described previously (Section 3.2). 

While generating aligned audio-visual event pairs would ideally provide richer temporal information for reasoning, implementing this step within the agent presents practical challenges. Unlike the data curation pipeline, the agent lacks access to a highly capable OLM like Gemini 2.5 Flash for precise event alignment. Furthermore, providing a large number of fine-grained event pairs may overwhelm the context capacity or reasoning capabilities of the agent's LLM. Therefore, we adopt a \textbf{targeted approach}: First, we prompt Qwen2.5-14B-Instruct with the visual and audio annotations, the question, and its choices. The model's task is to identify a list of specific \textbf{visual events} whose temporal localization is deemed necessary to answer the question correctly. Subsequently, we utilize Qwen2.5-VL-7B, functioning as a video temporal grounding model, to determine the start and end timestamps for each of these identified critical events. If the duration of a grounded event interval falls below a predefined threshold, the agent classifies this interval as critical. It then retrieves descriptions of both visual and audio events occurring within this specific, brief period using the previously mentioned approach, thereby establishing a localized alignment between concurrent events. Finally, we prompted Qwen2.5-14B-Instruct to determine the correct answer by providing it with the question, choices, the previously generated visual and audio annotations, and the extracted aligned event pairs.

\section{Experiment}
In this section, we present a comprehensive evaluation of recent Multimodal Large Language Models (MLLMs) on our benchmark to delineate their capability boundaries. Furthermore, we conduct ablation studies to investigate the key factors influencing model performance.
\subsection{Settings}
Our evaluation encompassed three distinct types of models. Firstly, we examine \textbf{OLMs}, including several open-source contenders like VideoLLaMA 2 \citep{cheng_videollama_2024}, Unified-IO 2 \citep{lu2023unifiedio2scalingautoregressive}, Qwen2.5-Omni \citep{xu_qwen25-omni_2025}, Qwen3-Omni \cite{Qwen3-Omni}, Ola \citep{liu_ola_2025} and our \textbf{Daily-Omni Agent}, as well as the proprietary Gemini 2.0 and 2.5 models \citep{noauthor_introducing_2024,gemini25}. Secondly, we assess \textbf{VLMs} including Qwen2.5-VL \citep{bai_qwen25-vl_2025} and GPT-4o \citep{openai_gpt-4o_2024} and \textbf{ALMs} including Audio Flamingo 3 \citep{goel2025audioflamingo3advancing} and Qwen2-Audio \citep{chu2024qwen2audiotechnicalreport}. We also evaluate the performance of some OLMs when provided with only visual inputs. Finally, we test some \textbf{LLMs} such as Deepseek-V3 \citep{deepseekai2025deepseekv3technicalreport}, GPT-4o \citep{openai_gpt-4o_2024} and Qwen2.5-14B\citep{qwen2025qwen25technicalreport} without visual and audio inputs to \textbf{evaluate the extent of textual bias} in our questions and choices. All model evaluations are conducted strictly according to instructions from their respective official repositories and accompanying documentation (such as "cookbooks" or developer guides).
\subsection{Main Results}

\begin{table}[htbp]
\centering
\caption{\textbf{Performance comparison of MLLMs on Daily-Omni}. Boldface and underline indicate the top two performers. Subscripts on 'Avg' for visual-only OLMs show the performance drop from their audio-visual counterparts. Abbreviations: AV Align = audio-visual alignment, Comp. = comparative, Ctx. Und. = context understanding, Evt. Seq. = event sequence, Infer. = inference, Reas. = reasoning, and 30s/60s denote duration-based subsets. Random guess accuracy is 25\%.
\textbf{Color code:} \colorbox{gray!20}{Gray background} indicates closed-source models and \colorbox{blue!10}{blue background} indicates open-source models.}
\scriptsize
\setlength{\tabcolsep}{2.5pt}

\resizebox{\linewidth}{!}{
\begin{tabular}{c | S[table-format=2.2] S[table-format=2.2] S[table-format=2.2] S[table-format=2.2] S[table-format=2.2] S[table-format=2.2] |S[table-format=2.2] S[table-format=2.2] |S[table-format=2.2]}
\toprule
\textbf{Methods} & \textbf{AV Align} & {\textbf{Comp.}} & \textbf{Ctx. Und.} & \textbf{Evt. Seq.} & {\textbf{Infer.}} & {\textbf{Reas.}}& {\textbf{30s}} & {\textbf{60s}} & {\textbf{Avg}}  \\
\midrule
\multicolumn{10}{c}{\textbf{Omni-Modal Language Models (With Visual and Audio)}} \\
\midrule
\rowcolor{blue!10} Qwen2.5-Omni-7B-Instruct & 48.32 & 69.47 & 58.55 & 58.17 & 76.62 & 73.14 & 64.61 & 59.09 & 62.07\\
\rowcolor{blue!10} Qwen2.5-Omni-3B-Instruct & 50.84 & 69.47 & 53.89 & 53.92 & 75.97 & 70.29 & 62.60 & 57.45 & 60.23\\
\rowcolor{blue!10} Qwen3-Omni-30B-A3B-Instruct & \underline{66.81} & \textbf{80.92} & 64.77 & 66.34 & \underline{81.17} & \underline{81.14} & \underline{71.87} & \underline{71.82} & 71.85\\
\rowcolor{blue!10} Qwen3-Omni-30B-A3B-Thinking & 65.97 & \textbf{80.92} & \underline{65.80} & \textbf{71.57} & \textbf{85.06} & 80.57 & \textbf{76.04} & 70.73 & \textbf{73.60}\\
\rowcolor{blue!10} Ola (7B) & 40.34 & 61.07 & 40.41 & 43.46 & 63.64 & 69.71 & 51.47 & 49.82 & 50.71\\
\rowcolor{blue!10} Unified-IO-2 L (1B)   & 27.31 & 22.90 & 26.42 & 27.78 & 29.87 & 29.14 & 27.67 & 27.09 & 27.40 \\
\rowcolor{blue!10} Unified-IO-2 XL (3B)   & 30.25 & 30.53 & 25.39 & 29.08 & 33.12 & 21.71 & 28.13 & 28.55 & 28.32  \\
\rowcolor{blue!10} Unified-IO-2 XXL (8B)  & 25.63 & 31.30 & 26.42 & 25.82 & 35.06 & 29.71 &26.74 & 30.00 & 28.24  \\
\rowcolor{blue!10} VideoLLaMA2 (7B) & 35.71 & 35.88 &35.75 & 31.70 & 40.91 & 34.29 & 38.02 & 31.82 & 35.17\\
\rowcolor{gray!20} Gemini 2.5 Flash & \textbf{73.82} & 66.41 & \textbf{72.04} & \underline{68.03} & 78.67 & \textbf{81.87} & 69.86 & \textbf{77.09} & \underline{73.06}\\
\rowcolor{gray!20} Gemini 2.5 Flash Lite & 57.56 & 68.70 & 56.48 & 52.61 & 79.22 & 69.71 & 63.52 & 60.00 & 61.90\\
\rowcolor{gray!20} Gemini 2.0 Flash & 62.18 & \underline{73.28} & 63.73 & 63.72 & 76.62 & 75.43 & 67.23 & 68.55 & 67.84\\
\rowcolor{gray!20} Gemini 2.0 Flash Lite & 55.04 & 64.89 & 58.03 & 54.25 & 74.03 & 72.00 & 62.44 & 60.00 & 61.32\\
\rowcolor{blue!10} \textbf{Daily-Omni-Baseline-Qwen2.5} & 51.68 & 68.70 & 60.10 & 53.92 & 78.57 & 71.43 & 63.99 & 59.27 & 61.82\\

\midrule
\multicolumn{10}{c}{\textbf{Omni-Modal Language Models (Visual Only)}} \\
\midrule
\rowcolor{blue!10} Qwen2.5-Omni-7B-Instruct & 34.45 & 58.78 & 47.67 & 49.67 & 62.99 & 54.86 & 48.69 & 51.09 & \multicolumn{1}{c}{$49.79_{\text{-}12.3}$} \\
\rowcolor{blue!10} Qwen2.5-Omni-3B-Instruct & 37.39 & 51.91 & 44.56 & 41.18 & 64.29 & 48.00 & 46.52 & 45.64 & \multicolumn{1}{c}{$46.12_{\text{-}14.1}$} \\
\rowcolor{blue!10} Qwen3-Omni-30B-A3B-Instruct & \textbf{47.90} & \textbf{67.18} & \textbf{56.48} & 53.92 & \textbf{70.78} & \underline{61.14} & \textbf{57.96} & \underline{57.64} & \multicolumn{1}{c}{$\textbf{57.81}_{\text{-}14.0}$} \\
\rowcolor{blue!10} Qwen3-Omni-30B-A3B-Thinking & \underline{44.96} & \underline{64.89} & \underline{55.96} & \textbf{59.48} & \underline{67.53} & 60.00 & 56.26 & \textbf{59.45} & \multicolumn{1}{c}{$\underline{57.73}_{\text{-}15.9}$} \\
\rowcolor{gray!20} Gemini 2.5 Flash & 37.55 & 37.21 & 40.43 & 44.78 & 57.05 & 53.29 & 42.35 & 47.46 & \multicolumn{1}{c}{44.61\ensuremath{_{\text{-}28.5}}} \\
\rowcolor{gray!20} Gemini 2.5 Flash Lite & 36.97 & 45.80 & 37.31 & 39.54 & 59.74 & 47.43 & 44.67 & 41.27 & \multicolumn{1}{c}{43.11\ensuremath{_{\text{-}18.8}}} \\
\rowcolor{gray!20} Gemini 2.0 Flash & 39.08 & 64.12 & \textbf{56.48} & \underline{56.21} & \underline{67.53} & \textbf{62.29} & \underline{56.57} & 55.45 & \multicolumn{1}{c}{56.06\ensuremath{_{\text{-}11.8}}} \\
\rowcolor{gray!20} Gemini 2.0 Flash Lite & 43.70 & 58.02 & 53.89 & 45.10 & 64.29 & 60.57 & 53.01 & 51.64 & \multicolumn{1}{c}{52.38\ensuremath{_{\text{-}8.9}}} \\

\midrule
\multicolumn{10}{c}{\textbf{Omni-Modal Language Models (Audio Only)}} \\
\midrule
\rowcolor{gray!20} Gemini 2.5 Flash & 46.64 & 55.73 & 44.56 & 42.48 & 70.78 & \textbf{78.86} & 55.64 & 52.18 & \multicolumn{1}{c}{54.05\ensuremath{_{\text{-}19.0}}} \\
\rowcolor{gray!20} Gemini 2.5 Flash Lite & 42.02 & 61.83 & 41.97 & 45.10 & 68.83 & 65.14 & 54.25 & 48.91 & \multicolumn{1}{c}{51.80\ensuremath{_{\text{-}10.1}}} \\
\rowcolor{blue!10} Qwen3-Omni-30B-A3B-Instruct & \underline{54.20} & \textbf{69.47} & \textbf{51.81} & \textbf{51.63} & \underline{74.03} & \textbf{78.86} & \underline{63.37} & \textbf{58.18} & \multicolumn{1}{c}{\textbf{60.99}\ensuremath{_{\text{-}10.9}}} \\
\rowcolor{blue!10} Qwen3-Omni-30B-A3B-Thinking & \textbf{54.62} & \underline{67.94} & \underline{49.22} & \underline{51.31} & \textbf{77.27} & \underline{77.71} & \textbf{65.22} & \underline{55.27} & \multicolumn{1}{c}{\underline{60.65}\ensuremath{_{\text{-}13.0}}} \\

\midrule
\multicolumn{10}{c}{\textbf{Visual Language Models (Visual Only)}} \\
\midrule
\rowcolor{gray!20} GPT-4o & 47.90 & 62.60 & 52.33 & 52.61 & {66.23} & {66.29} & {55.64} & {57.45} & {56.47}\\
\rowcolor{blue!10} Qwen2.5-VL-7B-Instruct & {36.97} &{46.56} & 33.68 & {37.91} & {51.95} & {44.00} & {39.26} & {42.36} & {40.68}\\
\rowcolor{blue!10} Qwen2.5-VL-3B-Instruct & 35.71 & 43.51 & {34.72} & 33.66 & 43.51 & 39.43 & 37.71 & 37.09 & 37.43\\
\rowcolor{blue!10} Qwen3-VL-4B-Instruct & 43.70 & 61.07 & 54.40 & 53.27 & 68.18 & 58.86 & 54.40 & 56.00 & 55.14\\
\rowcolor{blue!10} Qwen3-VL-8B-Instruct & 44.54 & 63.36 & 50.78 & 59.80 & 69.48 & 58.86 & 56.41 & 57.27 & 56.81\\
\rowcolor{blue!10} Qwen3-VL-30B-A3B-Instruct & 47.48 & 68.70 & 52.33 & 55.88 & 67.53 & 61.14 & 57.34 & 57.27 & 57.31\\

\midrule
\multicolumn{10}{c}{\textbf{Audio Language Models (Audio Only)}} \\
\midrule
\rowcolor{blue!10} Audio Flamingo 3 (7B) & 40.76 & 55.73 &  43.01 & 40.52 &65.58  & 68.00 & 50.23 & 49.45  & 49.87 \\
\rowcolor{blue!10} Qwen2-Audio (7B) & 28.99 &35.88 & 27.46 & 32.03 &33.77 &33.14 &31.22  &31.82 & 31.50 \\

\midrule
\multicolumn{10}{c}{\textbf{Textual Language Models (Without Visual and Audio)}} \\
\midrule
\rowcolor{gray!20} GPT-4o & 33.19 & 43.51 & 28.50 & 30.39 & 44.81 & 46.86 & 36.48 & 36.18 & 36.34\\
\rowcolor{gray!20} Deepseek-V3 (671B) & 31.93 & 41.22 & 29.02 & 29.41 & 44.81 & 46.29 & 35.24 &  36.00 & 35.59\\
\rowcolor{blue!10} Qwen2.5-14B-Instruct & 30.25 & 39.69 & 27.98& 28.43& 42.21& 42.86& 32.15 & 35.82& 33.83\\
\bottomrule
\end{tabular}}
\label{tab:mllm_comparison} 
\end{table}
\textbf{Table \ref{tab:mllm_comparison}} presents comprehensive evaluation results, shedding light on the real-world audio-visual understanding capabilities of contemporary MLLMs and our diagnostic reference baseline, the Daily-Omni Agent.
Firstly, earlier Omni-modal Language Models (OLMs) such as Unified-IO 2 and VideoLLaMA 2 demonstrate limited performance on our benchmark. Notably, their results are, in several instances, even surpassed by text-only LLMs. Furthermore, the Unified-IO 2 series displays a perplexing degradation in performance with increasing model size, an observation consistent with findings from OmniBench \citep{li_omnibench_2024}. A plausible explanation is that these models may not reliably synthesize cross-modal information under our evaluation setting.

Secondly, we observe a clear performance divide among recent OLMs, highlighting the importance of cross-modal temporal alignment for audio-visual QA. Our training-free Daily-Omni Agent, which makes temporally localized evidence more explicit via a decoupled pipeline, serves as a useful \emph{reference point} with an overall accuracy of 61.82\%. Models that perform substantially below this reference (such as Ola and earlier OLMs) appear to struggle with synchronizing parallel audio and visual streams for alignment-sensitive questions.
In contrast, models that achieve higher performance demonstrate stronger proficiency in audio-visual reasoning. While proprietary models like Gemini~2.5~Flash achieve top-tier performance (up to 73.06\%), among open-source models, the Qwen2.5-Omni and Qwen3-Omni series stand out. Their gains likely reflect a combination of factors, including training data and objectives, architectural choices for multimodal fusion, and temporal modeling components. In particular, explicit cross-modal positional designs (e.g., TMRoPE) may be one contributing technique for improving temporal awareness across modalities. Overall, these results suggest that robust spatio-temporal alignment mechanisms remain an important ingredient for tackling complex audio-visual reasoning tasks.

These results reveal that: (1) Advanced OLMs with stronger temporal handling (e.g., the Qwen Omni series) or proprietary systems (e.g., Gemini) demonstrate solid audio-visual understanding, though they still face challenges in complex temporal reasoning. (2) Models with weaker temporal integration tend to perform poorly on alignment-sensitive questions, even when provided with both audio and visual inputs. (3) Our QA construction and filtering pipeline produces challenging evaluation instances that probe the temporal reasoning boundaries of current MLLMs.

\begin{figure}[t]
    \centering
    \includegraphics[width=1\linewidth]{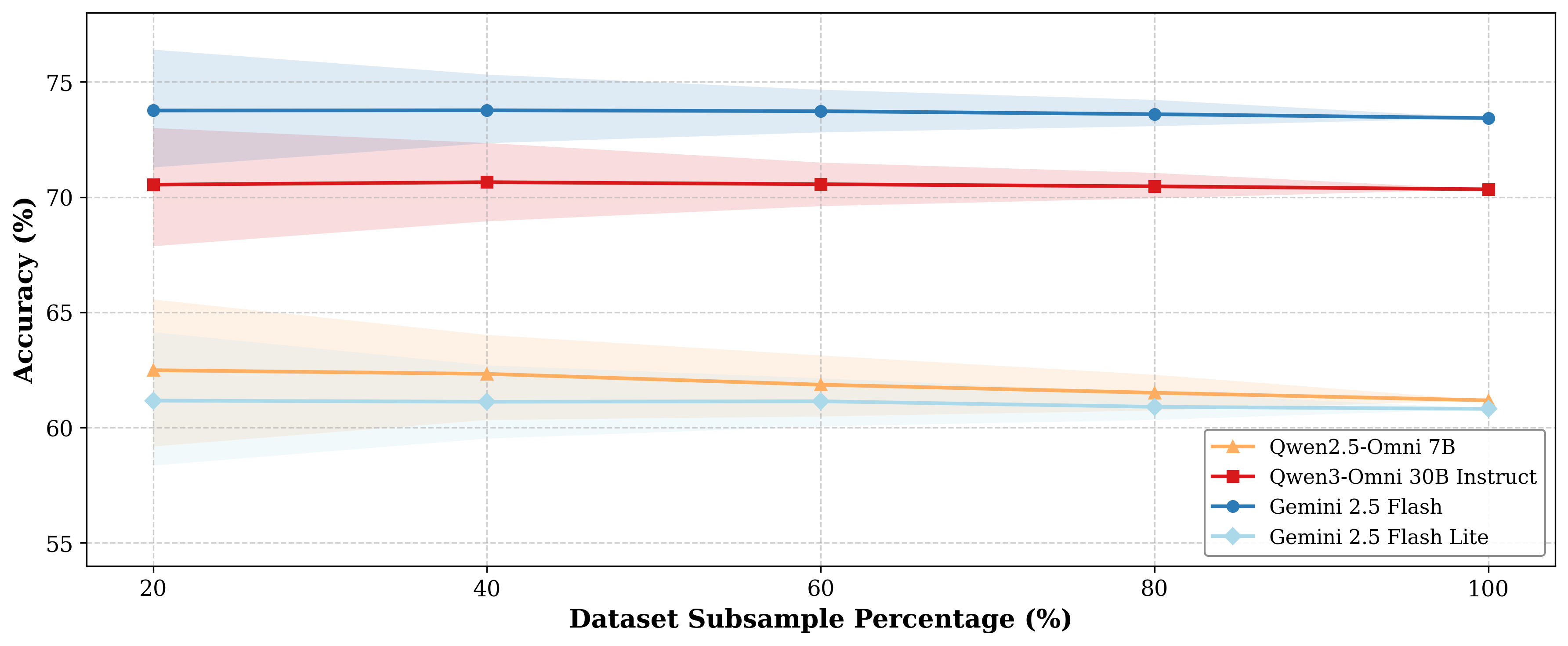}
    \caption{\textbf{Evaluation stability under video-level subsampling.} Lines show median overall accuracy and shaded bands denote the 5th--95th percentile across 200 trials.}
    \label{fig:scale_robustness}
\end{figure}

\subsection{Evaluation Stability via Subsampling}
To assess the stability of aggregate evaluation on Daily-Omni, we conduct a video-level subsampling study on representative OLMs. We repeatedly sample \{20,40,60,80\}\% of videos without replacement and evaluate overall accuracy on the induced QA subsets (Figure~\ref{fig:scale_robustness}). The estimates remain consistent across subsamples and the variability decreases rapidly as more videos are included: at 80\% of videos (approximately 958 questions), the 5th--95th percentile range is within \textbf{1.1--1.2 percentage points} for Gemini~2.5~Flash, Gemini~2.5~Flash~Lite, and Qwen3-Omni-30B. Overall, this subsampling analysis supports that Daily-Omni provides sufficiently stable aggregate evaluation to serve as a robust benchmark.

\subsection{Ablation Study}
To further validate the design of our proposed baseline and the intrinsic multimodal requirements of the Daily-Omni benchmark, we conduct comprehensive ablation studies focusing on temporal alignment methods and modality dependency.
\begin{figure}[t]
    \centering
    \begin{subfigure}[b]{0.57\linewidth}
        \centering
        \includegraphics[width=\linewidth]{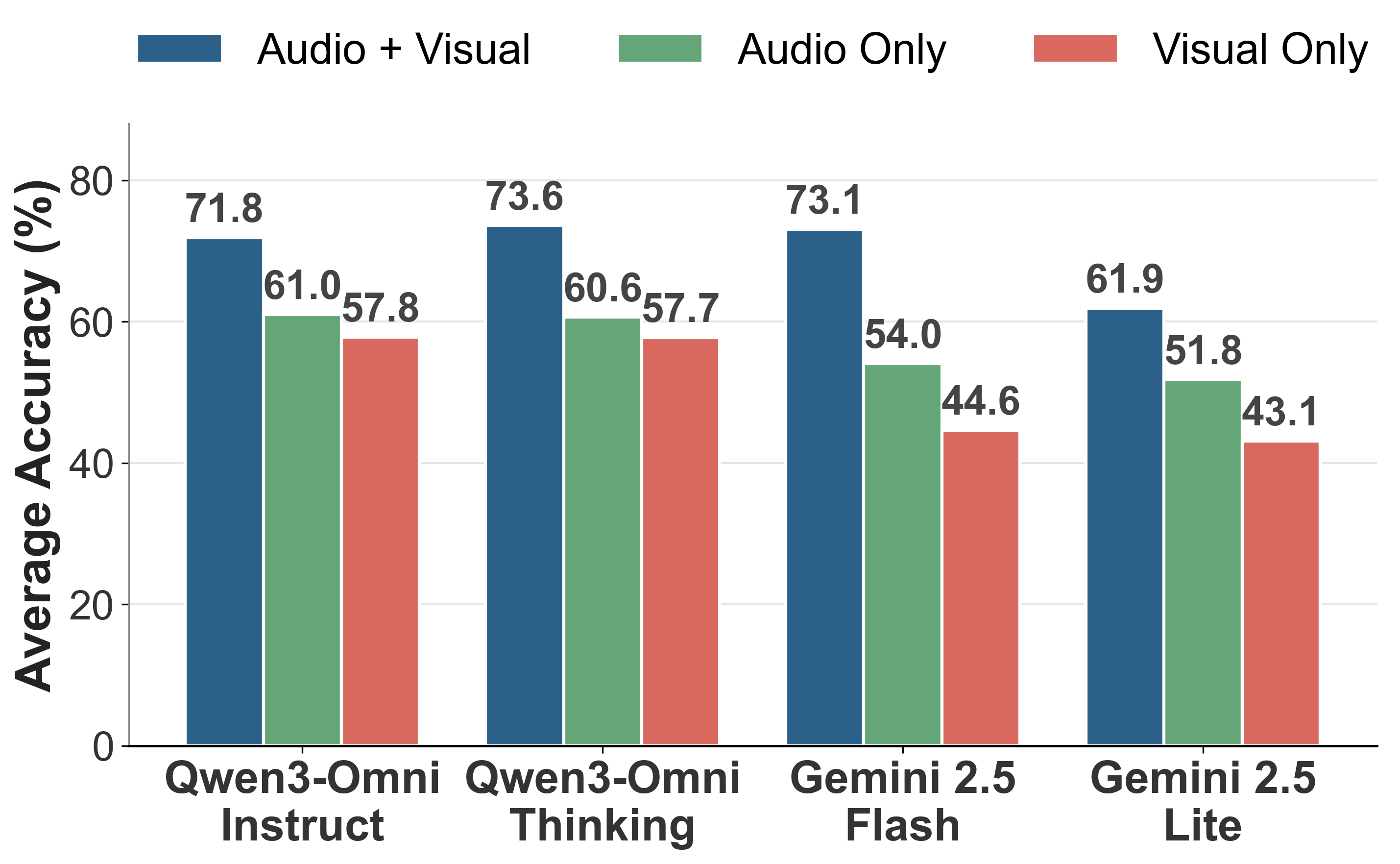}
        \caption{Modality ablation on Daily-Omni}
        \label{fig:sub_modality}
    \end{subfigure}
    \hfill
    \begin{subfigure}[b]{0.42\linewidth}
        \centering
        \includegraphics[width=\linewidth]{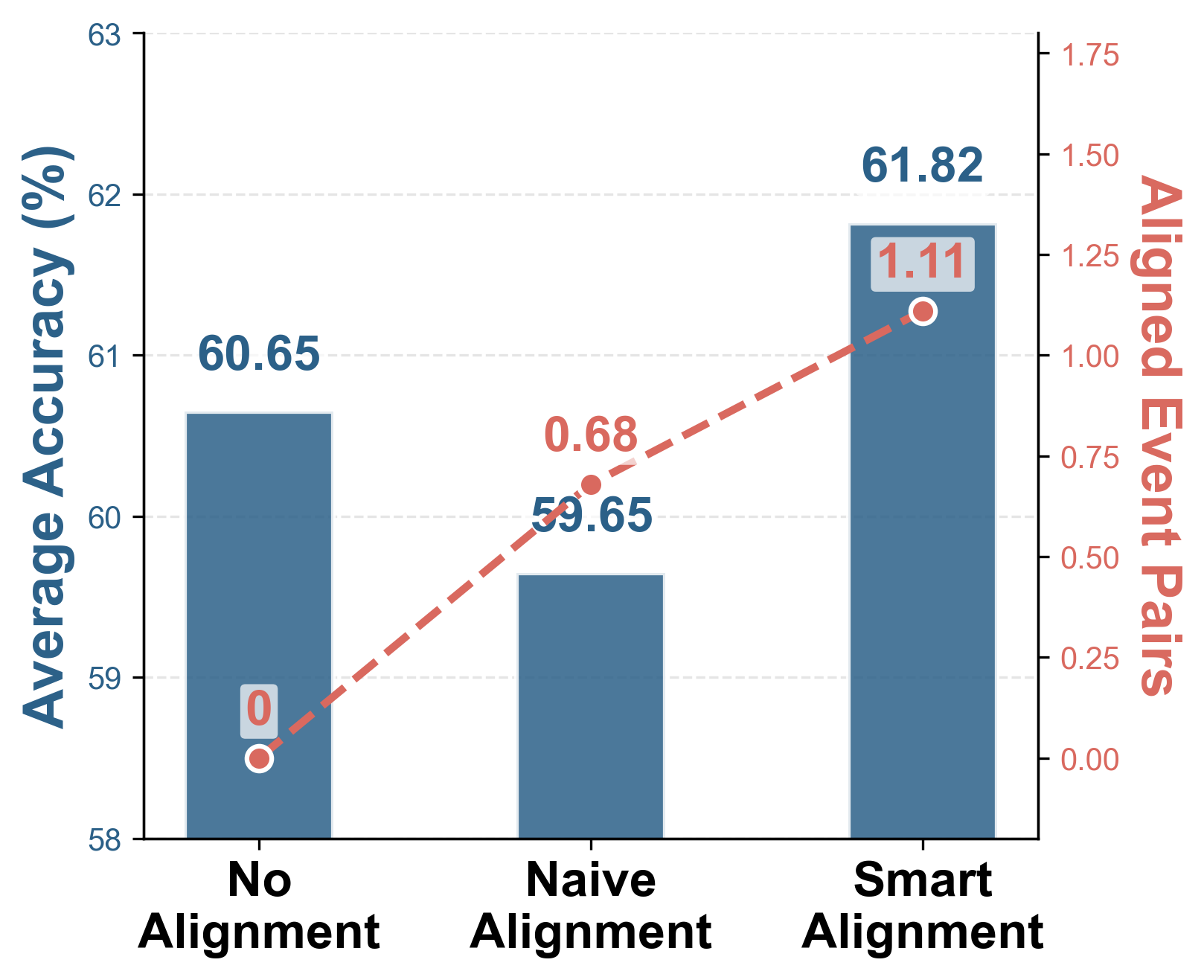}
        \caption{Impact of alignment methods}
        \label{fig:sub_alignment}
    \end{subfigure}  
    
    \caption{\textbf{Ablation studies validating the design of the benchmark and the baseline agent.} (a) Performance comparison of representative OLMs under full modalities (A+V), Audio-only, and Visual-only settings. (b) Comparison of average accuracy (bars) and aligned event pairs per question (line) using our baseline agent.}
    \label{fig:ablation_overview}
\end{figure}

\subsubsection{Modality Ablation}
To verify whether the tasks in Daily-Omni genuinely require the integration of both auditory and visual information, we evaluated representative top-performing OLMs (Qwen3-Omni-30B-A3B-Instruct/Thinking, Gemini 2.5 Flash, and Gemini 2.5 Flash Lite) under three conditions: full modalities (Audio + Visual), audio-only, and visual-only. The comparative results are illustrated in \textbf{Figure \ref{fig:ablation_overview}(a)}.

Upon ablating either modality, all evaluated OLMs exhibit a substantial decline in performance. For instance, the highly capable Gemini 2.5 Flash experiences a drastic accuracy drop from 73.06\% to 54.05\% without visual input, and down to 44.61\% without audio input. Similarly, the Qwen3-Omni models suffer a 13--16\% performance degradation when deprived of either modality. This significant performance gap between the full-modality and uni-modal settings strongly validates that the questions in our benchmark cannot be reliably answered using a single sensory channel or text-based guessing.

Interestingly, we observe that for these models, the audio-only performance is generally higher than the visual-only performance on our dataset. This observation highlights that unlike many existing video QA benchmarks that exhibit a strong visual bias (where audio is often redundant), Daily-Omni contains rich, indispensable acoustic events and spoken dialogues. In conclusion, this ablation confirms that achieving high performance on Daily-Omni strictly necessitates genuine multi-modal synergy and precise cross-modal integration.

\subsubsection{Daily-Omni Agent Alignment}
Our heuristic Daily-Omni Agent, leveraging Qwen2.5-VL-7B, Qwen2-Audio (7B), and Qwen2.5-Instruct-14B, achieves 61.82\% overall accuracy. The fact that this simple, decoupled pipeline surpasses smaller proprietary MLLMs and several unified open-source models strongly validates the core motivation of our benchmark: temporal alignment is the primary bottleneck in audio-visual reasoning.

To further study the effect of event alignment, we conducted an ablation study evaluating the Daily-Omni Agent's performance under three alignment scenarios:
(1) \textbf{No Alignment}: Generating comprehensive visual and audio captions as a base, but omitting any form of event alignment.
(2) \textbf{Naive Alignment}: Generating comprehensive visual and audio captions, then aligning events in a question-agnostic manner. Specifically, we prompt the VLM to enumerate visual events from the captions and localize each event by predicting its temporal boundaries (start/end). For each localized interval, we retrieve the corresponding audio events within the same time range using the ALM, and form aligned audio--visual event pairs.
(3) \textbf{Smart Alignment}: Generating comprehensive visual and audio captions and aligned events as stated in Section \textbf{3.3}. Note that for (2) and (3), we only consider generating an align event pair when the provided segment has a duration below a certain threshold to prevent confusion. 

The overall accuracy and the average number of aligned event pairs identified per question for each method are presented in \textbf{Figure \ref{fig:ablation_overview}(b)}. As expected, the Smart Alignment method achieves the highest average accuracy. This demonstrates that explicitly identifying and aligning relevant events significantly boosts the model's overall performance compared to simply generating global captions. Conversely, the Naive Alignment method exhibited a slight decline in accuracy relative to the No Alignment baseline. Careful analysis of the generated events and their temporal boundaries suggests that this outcome is likely attributable to the limitations of the Qwen2.5-VL-7B model. It appears that this model is not sufficiently powerful to reliably identify and retrieve the target event through a single query. Moreover, the temporal grounding process itself frequently yields imprecise results, leading to erroneous alignment. This issue with temporal grounding affects even the Smart Alignment method, occasionally producing incorrect or confusing aligned event pairs. Consequently, we hypothesize that equipping the agent with a more powerful open-vocabulary video temporal grounding model would unlock further significant improvements in its performance, mitigating the impact of imprecise temporal grounding.

\section{Conclusion}
This paper introduced Daily-Omni, an Audio-Visual Question Answering benchmark for evaluating temporally aligned multimodal reasoning in daily-life scenarios, together with a scalable QA construction and verification pipeline. Our evaluation of 24 foundation models under 37 model--modality settings shows that while recent MLLMs exhibit strong unimodal capabilities, they still struggle on questions that require fine-grained cross-modal temporal alignment. We also include a training-free modular reference baseline (Daily-Omni Agent) to help interpret alignment-sensitive failure modes and to illustrate the impact of making temporally localized evidence explicit. Overall, Daily-Omni highlights temporal alignment as a key bottleneck for current unified architectures. Future work should prioritize more accurate and robust multimodal temporal grounding and alignment mechanisms for audio-visual reasoning.

\bibliographystyle{plainnat}
\bibliography{main}

\end{document}